\documentclass[a4paper,fleqn]{cas-sc}

\usepackage{multirow}
\usepackage[numbers,square,sort&compress]{natbib}

\usepackage[ruled]{algorithm2e}
\usepackage{booktabs}
\usepackage[section]{placeins}
\ExplSyntaxOn
\cs_set:Npn \__first_footerline: {}
\ExplSyntaxOff
\def\tsc#1{\csdef{#1}{\textsc{\lowercase{#1}}\xspace}}
\tsc{WGM}
\tsc{QE}
\tsc{EP}
\tsc{PMS}
\tsc{BEC}
\tsc{DE}
\begin{document}
\let\WriteBookmarks\relax
\def\floatpagepagefraction{1}
\def\textpagefraction{.001}
\shorttitle{}
\shortauthors{}
%\begin{frontmatter}

\title [mode = title]{Intra-Class Diversity Loss for Person Re-Identification}

\author[1]{Wei Zhou}
\cormark[1]

\address[1]{Department of Artificial Intelligence, Guangzhou Huashang College, Guangzhou, 511300, China}

\author[1]{BaXian Chen}

\cortext[cor1]{Corresponding author}

\nonumnote{E-mail address: zhouwei32100@gmail.com (W. Zhou).
  }

\begin{abstract}
Person re-identification (ReID) aims to retrieve images of the same pedestrian across different cameras. However, the non-overlapping identities between training and testing sets, together with large intra-class variation, make it challenging to learn generalizable pedestrian representations. Most existing ReID loss functions mainly focus on intra-class compactness and inter-class separation, which may overlook useful intra-class differences. To address these issues, we propose a novel Intra-Class Diversity (ICD) loss that aims to maximize intra-class differences in classification-irrelevant directions. To the best of our knowledge, we are the first to introduce this class-weight-orthogonal diversity learning objective for person ReID. Specifically, based on the geometric relationship between the learned representation and its corresponding class-weight vector, we decompose each representation into the component along the class-weight direction and its orthogonal component. This decomposition provides a clear and interpretable geometric basis for modeling intra-class diversity. ICD then encourages larger differences among the orthogonal components of samples belonging to the same identity, while the classification loss maintains identity discrimination. The proposed ICD loss can be incorporated into existing ReID frameworks as an additional regularization term without modifying the feature extraction network. Extensive experiments on three widely used ReID datasets demonstrate the effectiveness of ICD and its complementarity with existing metric learning losses.
\end{abstract}


\maketitle

\section{Introduction}

The goal of person re-identification (ReID) is to retrieve images of the same pedestrian as a given query image from a large number of pedestrian images captured by multiple cameras. This task is challenging due to various factors such as lighting conditions, pedestrian poses, background clutter, and partial occlusion existing in each image. These challenges become more severe in occluded person ReID, where different images of the same pedestrian may contain significantly different visible body parts \cite{Nguyen2024CrossViTReID}. Hence, learning discriminative and robust features is crucial for the person ReID task, especially within the framework of deep learning. The challenge of deep feature learning lies in designing more efficient deep learning networks and loss functions that are more suitable for person ReID.

\begin{figure}
	\centering
		\includegraphics[width=1.0\textwidth]{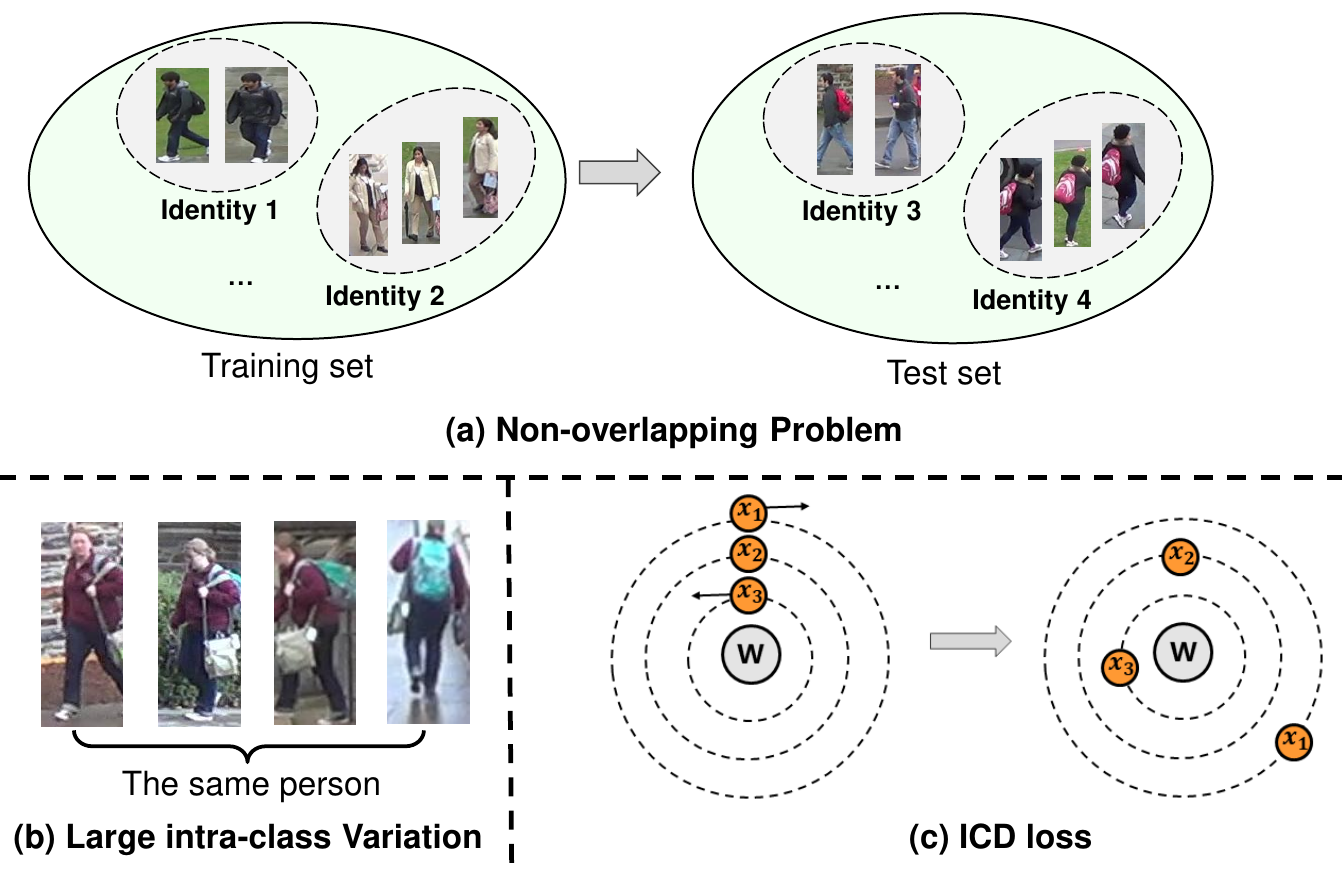}
\caption{Two specific challenges in person ReID and the motivation of the proposed ICD loss. (a) Non-overlapping identities between the training and testing sets. (b) Large intra-class variation among images of the same pedestrian. (c) Illustration of increasing intra-class differences in class-orthogonal directions, where $x_1$, $x_2$, and $x_3$ represent different images of the same identity, and $\mathbf{W}$ denotes the corresponding class-weight vector.}
	\label{fig:reid_motivation}
\end{figure}

For the design of loss functions, the common approach is to treat person re-identification as a classification problem and train the network using cross-entropy (CE) loss. And then, metric learning is employed to further enhance intra-class compactness and inter-class separation \cite{Luo2019BoT}. For example, the triplet loss \cite{Hermans2017Triplet} aims to pull samples of the same person class close to each other and push samples of different persons far away. The center loss \cite{Wen2016Center} tries to learn a class center for each person class and penalizes the distance of same-class samples to the class center. However, these methods do not fully consider two specific challenges in the person ReID task. That is, the class non-overlapping problem between the training and testing sets, as illustrated in Figure~\ref{fig:reid_motivation}(a), and the significant intra-class variation within different images of the same person, as shown in Figure~\ref{fig:reid_motivation}(b). Since the training and testing identities are non-overlapping, person ReID is not a standard closed-set classification task, and the learned representations need to generalize to unseen identities during testing \cite{Gu2022LFS}.

Recently, some works have become aware of the importance of intra-class variation in person ReID and proposed different solutions. For example, CPL \cite{Yang2022CPL} introduces a center prediction loss to allow more flexible intra-class distributions, while AdaSP \cite{Zhou2023AdaSP} proposes an adaptive positive mining strategy to handle different levels of intra-class variation. More recently, some studies have also explored representation diversity in visible-infrared and domain-generalizable person ReID \cite{Kim2024IRD,Jia2025DiverseReID,Li2026CDVM}. Although these methods have achieved promising results, they mainly address intra-class variation through sample selection, center prediction, or other feature learning strategies. However, these methods do not explicitly distinguish between the directions related to identity classification and the directions available for modeling intra-class variation. Therefore, how to maximize intra-class differences while maintaining identity discrimination remains an important problem. A more explicit and interpretable learning objective is desirable to better understand where and why intra-class differences should be encouraged.

Therefore, we also focus on intra-class variation in the person ReID task. Building upon existing research on classification, especially the geometric interpretation of angular classification \cite{Deng2019ArcFace}, we propose to enhance intra-class diversity in directions irrelevant to classification during training. Specifically, we refer to the directions orthogonal to the corresponding class-weight vector as classification-irrelevant directions. Based on this observation, we propose to maximize intra-class differences in classification-irrelevant directions, rather than simply encouraging all representations of the same identity to become more compact. Different from existing loss functions, our approach provides a clear geometric basis for modeling intra-class variation. In particular, the geometric relationship between the learned representation and its corresponding class-weight vector allows us to explicitly analyze why intra-class differences can be encouraged in these directions while preserving the alignment with the corresponding class-weight vector. In this way, we aim to address the class non-overlapping and large intra-class variation problems mentioned above, and further improve the ability of learned representations to generalize to unseen identities.

To summarize, in this paper, we propose a novel intra-class diversity loss for person ReID, named ICD loss. Different from conventional metric learning losses that mainly focus on the similarities between positive and negative samples, our ICD loss introduces a new regularization objective from the perspective of classification geometry. Specifically, we first map the learned representations and their corresponding class-weight vectors onto a unit hypersphere. Then, we decompose each representation into the component along the corresponding class-weight direction and the component orthogonal to that direction. Based on this geometric relationship, we propose to maximize the differences among representations of the same identity in classification-irrelevant directions, as illustrated in Figure~\ref{fig:reid_motivation}(c). Since ICD serves as an additional regularization term, it can be combined with existing classification and metric learning losses without modifying the original feature extraction network.

To verify the effectiveness of our proposed ICD loss, we conduct experiments on three widely used person ReID datasets, including Market1501 \cite{Zheng2015Market}, DukeMTMC-reID \cite{Ristani2016Duke}, and MSMT17 \cite{Wei2018MSMT}. We apply ICD to two representative Transformer-based feature learning networks, including ViT \cite{Dosovitskiy2021ViT} and TransReID \cite{He2021TransReID}. We also conduct ablation experiments to investigate the contribution of ICD and its complementarity with existing loss functions, such as triplet loss \cite{Hermans2017Triplet} and AdaSP \cite{Zhou2023AdaSP}. The experimental results show that our proposed ICD loss achieves consistent performance improvements on the three datasets.

Three contributions can be summarized as follows:

\begin{itemize}

\item We further investigate two specific challenges in person ReID, including the class non-overlapping problem between training and testing sets and the large intra-class variation among different images of the same person. To the best of our knowledge, we are the first to introduce a feature learning objective for person ReID that explicitly maximizes intra-class differences in classification-irrelevant directions.

\item We propose a novel intra-class diversity loss, named ICD loss, for person ReID. Different from existing metric learning losses, ICD is designed based on the geometric relationship between the learned representation vector and its corresponding class-weight vector. Our approach provides a clear geometric interpretation and theoretical analysis of how to encourage intra-class diversity in the class-orthogonal directions. Furthermore, ICD can serve as an additional regularization term without modifying the original feature extraction network.

\item Extensive experiments on three person ReID datasets demonstrate the effectiveness of the proposed ICD loss. The proposed loss consistently improves the performance of both ViT and TransReID networks. Ablation experiments and feature visualization demonstrate the complementarity of ICD with existing metric learning losses and provide additional evidence for its geometric motivation.

\end{itemize}

The rest part of this paper is organized as follows. Section 2 introduces the related works on occluded person ReID, deep metric learning, and the class non-overlapping and intra-class variation problems. Section 3 presents the proposed ICD loss and its theoretical analysis. The extensive experiments are conducted and the results are discussed in Section 4. Finally, the conclusion is drawn in Section 5.

\section{Related Work}

\subsection{Occluded Person ReID}

Person re-identification (ReID) aims to retrieve, from massive surveillance data, images of the same pedestrian given a query. This task is highly challenging due to the scale of the gallery and the substantial variations in illumination, contrast, pose, and other factors that affect each pedestrian image. In real-world deployments, pedestrians in either the query or gallery are frequently partially occluded, which reduces the amount of visible information and introduces noise from occluded regions. Consequently, ReID under occlusion is even more difficult.

A typical ReID system consists of two components: feature representation learning and deep metric learning. For feature representation under occlusion, existing approaches can be grouped into four categories. \textbf{(1) Local feature alignment.} These methods jointly learn global body features with local part features to obtain more discriminative representations \cite{Sun2018PCB,Li2021DPD}. For example, PCB \cite{Sun2018PCB} horizontally partitions a pedestrian image into several stripes and feeds them into part-specific classifiers to extract stronger local features. \textbf{(2) Pose-estimation-based alignment.} Still following the local-alignment idea, this line leverages an extra pre-trained pose estimator to align different body parts, and then aggregates part features as the final representation \cite{Gao2020PVPM}. \textbf{(3) Learning from visible parts.} The goal is to rely on visible regions while suppressing the influence of occluded areas. For instance, some methods use graph relations to improve local part features \cite{Wang2020HO}, while OAT \cite{Li2024OAT} uses second-order attention and uncertainty-guided feature fusion to reduce the influence of occlusion. \textbf{(4) Reconstruction of occluded regions.} Some methods reconstruct missing image regions or features to improve recognition performance under occlusion. For example, FRCE \cite{Zhang2024FRCE} introduces feature reconstruction and cross-enhancement to learn more reliable pedestrian representations.

Recently, Transformer-based methods such as CrossViT-ReID \cite{Nguyen2024CrossViTReID} have also been proposed to handle occlusion and appearance changes. However, most of the above methods mainly focus on improving feature extraction and try to learn \textbf{``tighter intra-class compactness and larger inter-class separation.''} This objective can be at odds with the substantial intra-class variability inherently present in occluded ReID.

\subsection{Deep Metric Learning for ReID}

Loss design is central to person ReID because, after extracting deep image representations, we expect to learn an embedding space in which samples of the same identity are close while those of different identities are well separated. Broadly, ReID losses fall into two categories according to their main objectives.

\textbf{Inter-class losses.} These objectives enlarge margins between identities and shape clear decision boundaries. The standard ID (cross-entropy) loss treats each person as a class. Margin-based Softmax variants (e.g., ArcFace/CosFace/AM-Softmax) \cite{Deng2019ArcFace,Wang2018CosFace,Wang2018AMSoftmax} further enforce angular or cosine margins to tighten inter-class separation. Batch-based formulations such as Triplet \cite{Hermans2017Triplet} and Circle \cite{Sun2020CircleLoss} loss also encourage larger inter-class distances while pulling positive samples together. For example, Triplet loss pulls positives and pushes negatives by a margin, while Circle loss optimizes pairwise similarities with different weights.

\textbf{Intra-class losses.} Most intra-class objectives drive compactness, aiming to collapse all images of the same person toward a single point (e.g., Center Loss \cite{Wen2016Center}). However, over-compactness compresses the representation space available to model legitimate variations within an identity: it overlooks the large instance-level differences induced by occlusion and can harm generalization to unseen IDs at test time. A few methods explicitly try to preserve intra-class diversity. For example, RLL \cite{Wang2019RLL} constrains positives within a margin band, while CPL \cite{Yang2022CPL} predicts the leave-one-out class center to allow more flexible intra-class distributions. More recently, AdaSP \cite{Zhou2023AdaSP} introduces adaptive positive mining to handle different levels of intra-class variation, while IRD \cite{Kim2024IRD} explores diverse intra-identity representations for visible-infrared ReID.

However, these methods mainly model intra-class variation through sample relationships or distribution constraints, without explicitly identifying the directions in which intra-class differences should be encouraged. Inspired by ArcFace, we address this issue by mapping features onto a hypersphere during training and maximizing same-identity differences along directions orthogonal to the corresponding class-weight vector. This provides a clear geometric interpretation of intra-class diversity. Our loss is naturally compatible with standard classification and metric learning objectives and can serve as a complementary term to improve generalization to unseen identities.

\subsection{Non-Overlapping IDs and Intra-class Variation in Occluded ReID}

In ReID, \textbf{train--test identity non-overlap} is the default setting. In practical deployments, people label identities from existing surveillance videos for training, and at test time submit a new query identity with the goal of retrieving all of its occurrences from the gallery images. This protocol naturally yields disjoint training and testing IDs. However, the implications of this problem for loss function design have received relatively limited attention. A common practice is to cast ReID as a standard image classification problem: train the feature extractor with cross-entropy, and further enforce intra-class compactness using Triplet loss.

Our motivation is to view ReID as an open-set retrieval task rather than closed-set classification. Because the model must transfer from training to unseen classes, preserving some intra-class flexibility may help enhance generalization. For instance, in few-shot learning, self-supervised knowledge distillation \cite{Rajasegaran2021SSKD} learns equivariant representations while preserving intra-class diversity. Likewise, in domain-adaptive classification, BMD \cite{Qu2022BMD} employs multi-center class prototypes to explicitly model within-class diversity. More recently, DiverseReID \cite{Jia2025DiverseReID} and CDVM \cite{Li2026CDVM} have also explored the benefits of modeling appearance and domain variations for generalizable person ReID.

For occluded ReID, such intra-class flexibility is especially critical. Because beyond being open-set, occlusions and pose changes lead instances of the same identity to expose non-overlapping visible parts, producing large within-class variance. Collapsing all images of one ID to a single point in feature space is therefore suboptimal---potentially harmful.

Accordingly, we propose an intra-class diversity loss for person ReID. During training, we map features onto a hypersphere and maximize same-identity differences along directions orthogonal to the corresponding class-weight vector, while using the classification loss to maintain identity discrimination. This design aims to address the class non-overlap between training and testing sets and the substantial intra-class variation in person ReID, improving generalization to unseen IDs.

\begin{figure}
	\centering
		\includegraphics[width=1.0\textwidth]{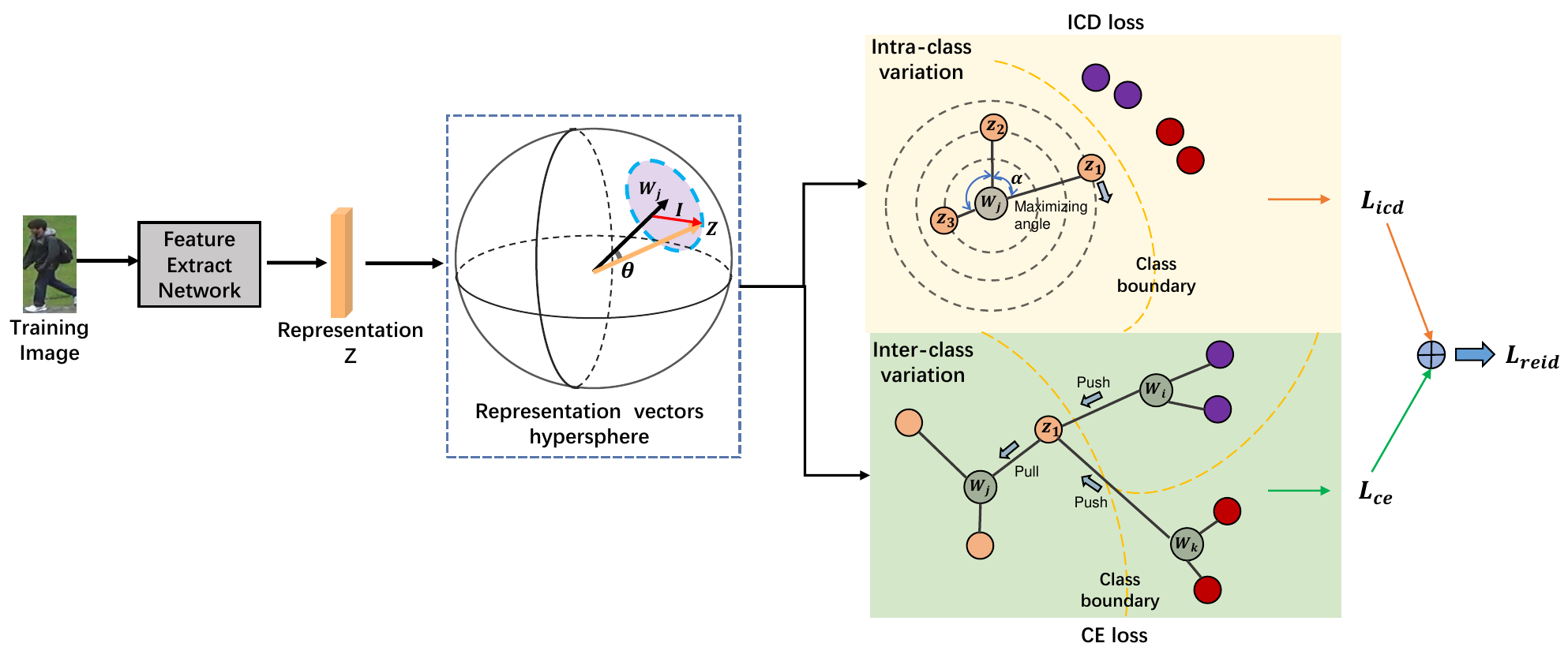}
\caption{The overall framework of the proposed ICD loss for person ReID. First, The feature learning network is used to extract the pedestrian representation $\mathbf{Z}$ from the input image.
Next, the pedestrian representation $\mathbf{Z}$ is projected onto a unit hypersphere together with the class weight vector $\mathbf{W_{j}}$ in the linear classifier.
Finally, the ICD loss and the CE loss are calculated to guide the training of the network together.}
	\label{fig:icd_framework}
\end{figure}

\section{Proposed Approach}
\label{sec:1}
In this section, we introduce the proposed intra-class diversity (ICD) loss for person re-identification in detail. First, we review why current person ReID methods usually treat ReID as an identity classification task and use the cross-entropy loss as the basic training objective. Then, we reformulate the classification problem from the perspective of the angle between the learned representation vector and its corresponding class-weight vector \cite{Deng2019ArcFace}. Based on this geometric view, we show that, when the angle between the representation and the class-weight vector remains unchanged, the representation can still vary in the class-orthogonal direction without changing the classification result. Therefore, we propose to preserve the intra-class diversity of the same person by maximizing the difference among samples in the class-orthogonal direction. Finally, we give the formal definition of the proposed ICD loss and present the overall training objective of the feature learning network.

\subsection{Overview}
\label{sec:icd_overview}

\indent\setlength{\parindent}{1em} The overall framework of our proposed ICD loss is shown in Figure 2. Given an input pedestrian image $\mathbf{X}\in\mathbb{R}^{H\times W\times C}$ with resolution $(H,W)$ and channel $C$, we first use a deep feature learning network to extract its representation vector $\mathbf{Z}\in\mathbb{R}^{d}$, where $d$ is the dimension of the learned representation. Then, the extracted representation is fed into an linear classification layer to calculate the classification scores and the cross-entropy (CE) loss. Similar to existing classification-based ReID methods, the CE loss is used to guide the network to learn discriminative representations for different pedestrian identities.

\indent\setlength{\parindent}{1em} Different from existing ReID loss functions that mainly focus on intra-class compactness and inter-class separation, our proposed ICD loss pays additional attention to intra-class variation. Specifically, we first apply $L2$ normalization to learn the representation vector $\mathbf{Z}$ and its corresponding class-weight vector $\mathbf{W_{j}}$, so that both vectors lie on a unit hypersphere.Based on the geometric relationship between these two vectors, we decompose the representation vector into two components, including the component along the corresponding class-weight direction and the component orthogonal to that direction. Next, we propose to measure the differences among different images of the same pedestrian in the class-orthogonal direction, and use these differences to construct the ICD loss. In this way, the ICD loss encourages the network to preserve intra-class diversity during training.

\indent\setlength{\parindent}{1em} Finally, the proposed ICD loss is combined with the conventional CE loss to train the feature learning network together. ICD is introduced as an additional regularization term during training. During testing, only the learned feature representations are used for identity matching, without calculating the ICD loss. The detailed formulation and implementation of the proposed ICD loss will be introduced in the following section.

\subsection{Intra-Class Diversity Loss for Person ReID}
\subsubsection{Review of the Classification Problem}
\label{sec:classification_review}
\begin{figure}
	\centering
		\includegraphics[width=0.85\textwidth]{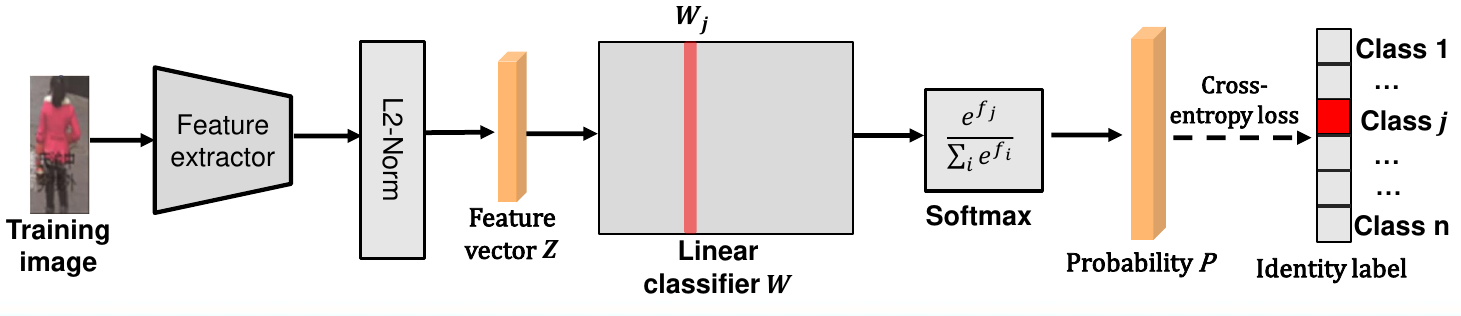}
\caption{The training pipeline of identity classification for person ReID. The feature learning network extracts the pedestrian representation, which is then fed into a linear classifier to calculate identity classification scores and the cross-entropy loss.}
	\label{fig:identity_classification}
\end{figure}

\indent\setlength{\parindent}{1em} Currently, most person ReID methods treat the ReID task as an identity classification problem during training, since the identity label of each training image is already known. Therefore, the cross-entropy (CE) loss is widely used as the basic training objective for learning discriminative pedestrian representations. As shown in Figure~\ref{fig:identity_classification}, given an input pedestrian image, we first use a deep feature learning network to extract its representation vector $\mathbf{Z}$ with $L2$ normalization. Then, the learned representation is fed into a linear classifier, followed by the Softmax function to predict the identity label of the input image. Finally, the CE loss is calculated to guide the training of the feature learning network.

\indent\setlength{\parindent}{1em} Specifically, let $\mathbf{Z}\in\mathbb{R}^{d}$ denote the learned representation vector, and let $\mathbf{W}\in\mathbb{R}^{C\times d}$ denote the weight matrix of the linear classifier, where $d$ is the dimension of the representation and $C$ is the number of training identities. For the $j$-th identity, the corresponding class-weight vector is denoted as $\mathbf{W}_j$. The classification score can be calculated by the inner product between the representation vector $\mathbf{Z}$ and the class-weight vector $\mathbf{W}_j$:

\begin{equation}
f_j = \mathbf{W}_j^{T}\mathbf{Z}
= \|\mathbf{W}_j\|\cdot\|\mathbf{Z}\|\cos\theta_j,
\end{equation}
where $\theta_j$ is the angle between $\mathbf{Z}$ and $\mathbf{W}_j$. Following the normalized classification setting, we apply L2 normalization to both vectors, so that $\|\mathbf{Z}\|=\|\mathbf{W}_j\|=1$. Therefore, the classification score can be further written as:
\begin{equation}
f_j = \mathbf{W}_j^{T}\mathbf{Z}
= \cos\theta_j.
\end{equation}
Based on the Softmax function, the predicted probability of the $j$-th identity is calculated as:
\begin{equation}
p_j = \frac{e^{s\cos\theta_j}}
{\sum_{k=1}^{C}e^{s\cos\theta_k}},
\end{equation}
where $s$ is a scaling factor. For a given image with ground-truth identity $y$, the CE loss can be defined as $\mathcal{L}_{ce}=-\log p_y$. Therefore, the classification problem can be understood from the perspective of angular similarity between the learned representation vector and the corresponding class-weight vector.

\begin{figure}
	\centering
		\includegraphics[width=0.85\textwidth]{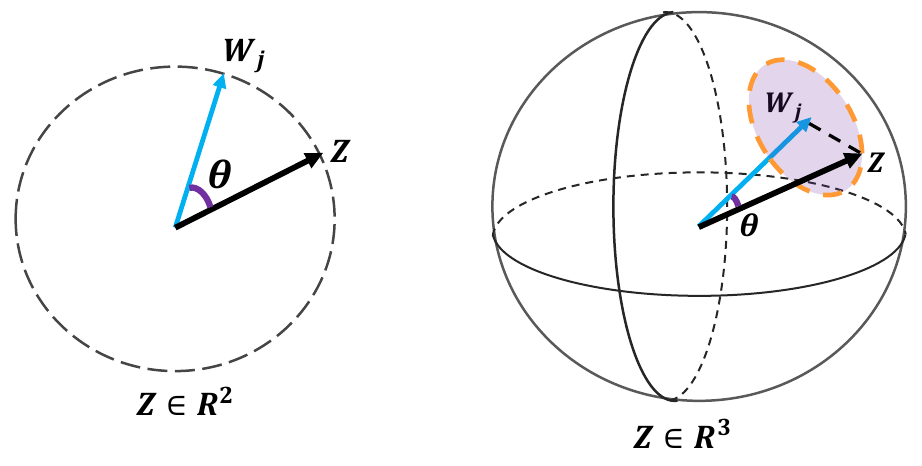}
\caption{Geometric interpretation of the classification problem. After L2 normalization, the representation vector and class-weight vector lie on a unit hypersphere. Their cosine similarity is determined by the angle between them.}
	\label{fig:classification_geometry}
\end{figure}

\indent\setlength{\parindent}{1em} As illustrated in Figure~\ref{fig:classification_geometry}, after L2 normalization, the representation vector $\mathbf{Z}$ and the class-weight vector $\mathbf{W}_j$ can be regarded as points on a unit hypersphere. From this geometric perspective, a smaller angle $\theta_j$ indicates a larger cosine similarity between these two vectors. More importantly, different representation vectors can have the same angle to the corresponding class-weight vector, although their directions on the hypersphere are different. In this case, the cosine similarity to the corresponding class-weight vector remains unchanged. Based on this observation, we further consider the variation of learned representations in directions orthogonal to the corresponding class-weight vector. In the following sections, we refer to these directions as classification-irrelevant directions and propose to maximize intra-class differences along these directions.
\subsubsection{Intra-Class Diversity Loss}
\indent\setlength{\parindent}{1em} As mentioned above, most existing person ReID loss functions mainly focus on increasing intra-class compactness and inter-class separation. However, due to the large intra-class variation and the non-overlapping identities between training and testing sets, forcing different representations of the same pedestrian to be highly compact may lead to overfitting on the training identities. Therefore, different from conventional metric learning methods, we propose an intra-class diversity (ICD) loss to preserve the intra-class variation of the learned representations during training. Specifically, we aim to maximize the differences among different images of the same pedestrian in classification-irrelevant directions, namely, the directions orthogonal to the corresponding class-weight vector, as illustrated in Figure~\ref{fig:reid_motivation}(c).

\indent\setlength{\parindent}{1em} From the geometric perspective, the learned representation vector can be decomposed into two components, including the component along its corresponding class-weight direction and the component orthogonal to that direction. As discussed in Section 3.2.1, after L2 normalization, the representation vector $\mathbf{Z}$ and its corresponding class-weight vector $\mathbf{W}_j$ lie on a unit hypersphere. Therefore, the component of $\mathbf{Z}$ in the class-orthogonal direction can be calculated as:

\begin{equation}
\mathbf{I}=\mathbf{Z}-\mathbf{W}_j\cos\theta,
\label{eq:orthogonal_component}
\end{equation}

where $\theta$ is the angle between the representation vector $\mathbf{z}$ and the class-weight vector $\mathbf{W}_j$, and $\mathbf{I}$ denotes the projection of $\mathbf{z}$ onto the direction orthogonal to $\mathbf{W}_j$. Based on this decomposition, we propose to measure the differences among different representations of the same pedestrian in the class-orthogonal direction.

Formally, suppose that there are $m$ pedestrian identities, and each identity contains $n$ different images. We use $\mathbf{Z}_{j,i}$ to denote the representation vector of the $i$-th image belonging to the $j$-th identity, and $\mathbf{W}_j$ to denote its corresponding class-weight vector. The orthogonal component of $\mathbf{Z}_{j,i}$ is denoted as $\mathbf{I}_{j,i}$. Then, the proposed intra-class diversity loss can be defined as follows:

\begin{equation}
\begin{aligned}
\mathcal{L}_{intra}
=&\sum_{j=1}^{m}
\sum_{i=1}^{n}
\sum_{\substack{k=1\\k\neq i}}^{n}
\left(
\frac{\mathbf{I}_{j,i}}{\|\mathbf{I}_{j,i}\|_2}
\cdot
\frac{\mathbf{I}_{j,k}}{\|\mathbf{I}_{j,k}\|_2}
\right)\\
=&\sum_{j=1}^{m}
\sum_{i=1}^{n}
\sum_{\substack{k=1\\k\neq i}}^{n}
\left(
\frac{\mathbf{Z}_{j,i}-\mathbf{W}_j\cos\theta_{j,i}}
{\|\mathbf{Z}_{j,i}-\mathbf{W}_j\cos\theta_{j,i}\|_2}
\cdot
\frac{\mathbf{Z}_{j,k}-\mathbf{W}_j\cos\theta_{j,k}}
{\|\mathbf{Z}_{j,k}-\mathbf{W}_j\cos\theta_{j,k}\|_2}
\right).
\end{aligned}
\label{eq:icd_loss}
\end{equation}

In Equation~\ref{eq:icd_loss}, the differences between representations of the same identity are measured by the cosine similarity of their normalized orthogonal components. A larger cosine similarity indicates that the two orthogonal components have more similar directions. Therefore, by minimizing the proposed ICD loss, we aim to maximize the intra-class differences of the learned representations in classification-irrelevant directions.

Finally, the proposed ICD loss is combined with the conventional cross-entropy loss to train the feature learning network together. The overall training objective can be written as:

\begin{equation}
\mathcal{L}_{reid}
=
\mathcal{L}_{ce}
+
\lambda\mathcal{L}_{intra},
\label{eq:overall_loss}
\end{equation}

where $\mathcal{L}_{ce}$ is the cross-entropy loss and $\lambda$ is a hyperparameter used to control the contribution of the proposed ICD loss. In this way, the CE loss mainly focuses on identity discrimination, while the ICD loss additionally encourages intra-class diversity during training. The geometric interpretation and theoretical analysis of the proposed loss will be discussed in the following section.
\subsubsection{Theoretical Analysis of the ICD Loss}
\begin{figure}
    \centering
    \includegraphics[width=\linewidth]{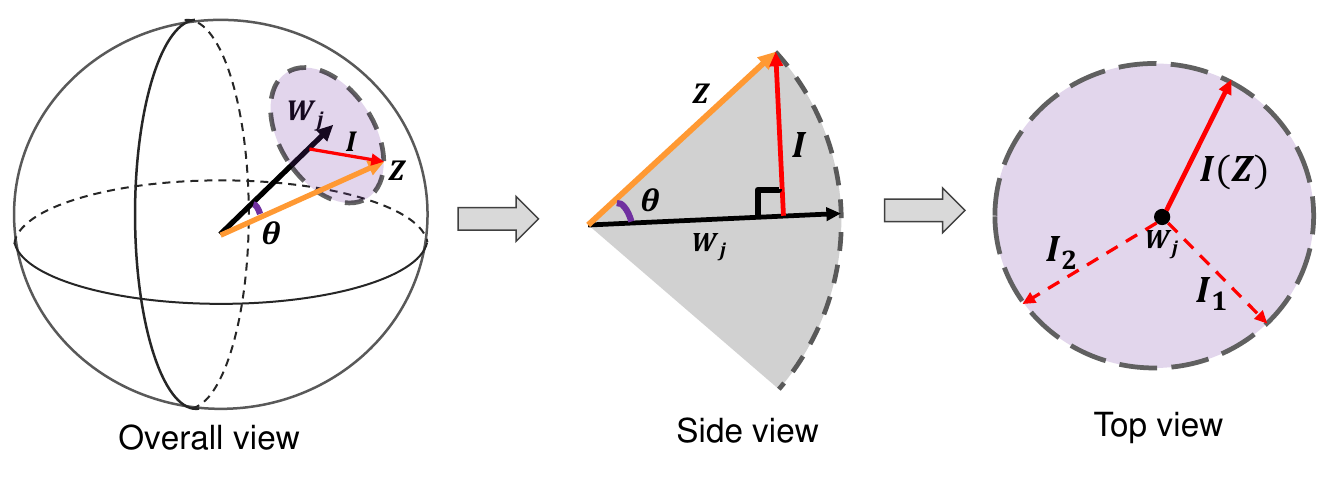}
    \caption{Geometric interpretation of the proposed ICD loss. The overall view shows the relationship between the representation vector and its corresponding class-weight vector. The side view illustrates the decomposition of the representation vector, while the top view shows the differences among the orthogonal components of different images belonging to the same identity.}
    \label{fig:icd_geometry}
\end{figure}

In this section, we further analyze the proposed ICD loss from the geometric perspective. As discussed in Section 3.2.2, our proposed ICD loss aims to maximize intra-class differences in classification-irrelevant directions. Specifically, we refer to the directions orthogonal to the corresponding class-weight vector as classification-irrelevant directions from the geometric perspective of identity classification. To better understand this idea, we first analyze the geometric relationship between the representation vector and its orthogonal component, as illustrated in Figure~\ref{fig:icd_geometry}.

Specifically, let $\mathbf{Z}$ denote a normalized representation vector and $\mathbf{W}_j$ denote its corresponding normalized class-weight vector. Following Equation~\ref{eq:orthogonal_component}, the orthogonal component is defined as $\mathbf{I}=\mathbf{Z}-\mathbf{W}_j\cos\theta$, where $\theta$ is the angle between $\mathbf{Z}$ and $\mathbf{W}_j$. Since both vectors have unit length, we can obtain:

\begin{equation}
\begin{aligned}
\mathbf{W}_j^T\mathbf{I}
&=\mathbf{W}_j^T
(\mathbf{Z}-\mathbf{W}_j\cos\theta)\\
&=\mathbf{W}_j^T\mathbf{Z}
-\mathbf{W}_j^T\mathbf{W}_j\cos\theta\\
&=\cos\theta-\cos\theta=0.
\end{aligned}
\label{eq:icd_orthogonality}
\end{equation}

Therefore, $\mathbf{I}$ is orthogonal to the corresponding class-weight vector $\mathbf{W}_j$. Furthermore, based on the geometric relationship shown in the side view of Figure~\ref{fig:icd_geometry}, the length of the orthogonal component can be calculated as:

\begin{equation}
\|\mathbf{I}\|_2
=\sqrt{1-\cos^2\theta}
=\sin\theta.
\label{eq:icd_orthogonal_norm}
\end{equation}

Based on the above analysis, when the angle $\theta$ remains unchanged, the length of the orthogonal component also remains unchanged. However, its direction can still vary in the space orthogonal to $\mathbf{W}_j$. As shown in the top view of Figure~\ref{fig:icd_geometry}, different representation vectors of the same identity can have different orthogonal directions while maintaining the same angle to their corresponding class-weight vector. In this case, their cosine similarities to $\mathbf{W}_j$ remain unchanged. Therefore, the class-orthogonal space provides additional directions for modeling intra-class variation.

Next, we further analyze the relationship between the orthogonal components and the similarity of different representations. Given two normalized representation vectors $\mathbf{Z}_{j,i}$ and $\mathbf{Z}_{j,k}$ belonging to the same identity, we can decompose them as:

\begin{equation}
\begin{aligned}
\mathbf{Z}_{j,i}
&=\mathbf{W}_j\cos\theta_{j,i}
+\mathbf{I}_{j,i},\\
\mathbf{Z}_{j,k}
&=\mathbf{W}_j\cos\theta_{j,k}
+\mathbf{I}_{j,k}.
\end{aligned}
\label{eq:icd_feature_decomposition}
\end{equation}

Since the orthogonal components are perpendicular to $\mathbf{W}_j$, the inner product between the two representations can be written as:

\begin{equation}
\begin{aligned}
\mathbf{Z}_{j,i}^T\mathbf{Z}_{j,k}
=&\cos\theta_{j,i}\cos\theta_{j,k}+\mathbf{I}_{j,i}^T\mathbf{I}_{j,k}.
\end{aligned}
\label{eq:icd_feature_similarity}
\end{equation}

Furthermore, let $\phi_{i,k}$ denote the angle between the two orthogonal components. Based on Equation~\ref{eq:icd_orthogonal_norm}, we have:

\begin{equation}
\begin{aligned}
\mathbf{Z}_{j,i}^T\mathbf{Z}_{j,k}
=&\cos\theta_{j,i}\cos\theta_{j,k}+\sin\theta_{j,i}\sin\theta_{j,k}
\cos\phi_{i,k}.
\end{aligned}
\label{eq:icd_angular_similarity}
\end{equation}

From Equation~\ref{eq:icd_angular_similarity}, it can be observed that, when the angles between the representations and their corresponding class-weight vector are fixed, decreasing $\cos\phi_{i,k}$ can reduce the similarity between representations of the same identity. Meanwhile, their cosine similarities to the corresponding class-weight vector remain unchanged. This provides a geometric explanation for the proposed ICD loss. By minimizing the cosine similarity between the orthogonal components, ICD encourages the learned representations to preserve more intra-class diversity.

Based on the above theoretical analysis, the learned representations of the same identity can have different directions in the class-orthogonal space while maintaining their cosine similarities to the corresponding class-weight vector. This provides a geometric explanation for our proposed ICD loss. Specifically, ICD aims to maximize intra-class differences in classification-irrelevant directions, while the CE loss is used to maintain identity discrimination during training. Here, classification-irrelevant directions refer to the directions orthogonal to the corresponding class-weight vector, rather than directions that are strictly irrelevant to all classification scores.

\subsubsection{Implementation of the ICD Loss}
\indent\setlength{\parindent}{1em} In this section, we introduce the implementation details of the proposed ICD loss. Different from existing methods that modify the feature extraction network, our ICD loss is implemented as an additional regularization term in the identity classification branch. Therefore, it can be directly integrated into existing ReID training frameworks without changing the original feature extraction network. The overall training procedure is summarized in Algorithm 1.

Specifically, given a mini-batch containing $B$ training images, we first use the feature learning network to extract their representation vectors. Then, we apply L2 normalization to the learned representations and the class-weight vectors in the linear classifier. In our implementation, we adopt an ArcFace-style classification layer to calculate the classification scores and the CE loss. Meanwhile, for each training image, we select its corresponding class-weight vector according to the identity label, and calculate the class-orthogonal component following Equation~\ref{eq:orthogonal_component}.

Next, we calculate the cosine similarity between the normalized orthogonal components of different images belonging to the same identity. As discussed in Section 3.2.2, the original ICD loss aims to maximize intra-class differences in classification-irrelevant directions. In practical implementation, we can adopt different strategies to calculate this regularization term. For example, instead of directly summing the similarities of all intra-class pairs, we can only select the pairs with positive cosine similarity and calculate their average similarity. In this way, the regularization term mainly penalizes intra-class pairs whose orthogonal components have similar directions, while avoiding continuously pushing apart pairs with non-positive similarity.

In addition, we detach the corresponding class-weight vectors and classification cosine values when calculating the orthogonal components. In this way, the ICD branch mainly updates the feature learning network, while the CE loss continues to optimize the identity classification branch. For numerical stability, a small constant is also used when normalizing the orthogonal components. Finally, the CE loss and ICD loss are combined to train the feature learning network together. The ICD loss is controlled by a weighting parameter $\lambda$, which determines its contribution to the overall training objective.
\begin{algorithm}[t]
\caption{Training a ReID network with the ICD loss}
\label{alg:icd_training}
\LinesNumbered

\KwIn{Training set $\mathcal{D}$; feature extractor $F$;
classifier $\mathbf{W}$; loss weight $\lambda$.}

Initialize $F$ and $\mathbf{W}$\;

\For{each training mini-batch}{

Sample $m$ identities with $n$ images per identity\;

Extract and normalize representations
$\mathbf{Z}_{j,i}=\mathrm{Norm}(F(x_{j,i}))$\;

Normalize class weights $\mathbf{W}_j$ and
calculate CE loss $\mathcal{L}_{ce}$\;

Compute orthogonal components
$\mathbf{I}_{j,i}
=\mathbf{Z}_{j,i}
-\mathbf{W}_j\cos\theta_{j,i}$\;

Calculate ICD loss
$\mathcal{L}_{intra}$ using
Equation~\ref{eq:icd_loss}\;

Compute total loss
$\mathcal{L}_{reid}
=\mathcal{L}_{ce}+\lambda\mathcal{L}_{intra}$\;

Update $F$ and $\mathbf{W}$
by back-propagation\;

}
\end{algorithm}

\FloatBarrier

\section{Experiments}

In this section, we conduct extensive experiments on three widely used person ReID datasets to evaluate the effectiveness of our proposed ICD loss. First, we introduce the datasets, evaluation metrics, and implementation details in Section 4.1. Then, we conduct ablation studies to analyze the contribution of the proposed ICD loss and its complementarity with existing training objectives in Section 4.2. After that, we compare our method with existing state-of-the-art ReID methods in Section 4.3. Finally, we provide t-SNE visualization to analyze the intra-class distribution of the learned representations in Section 4.4.

\subsection{Datasets and Implementation Details}

\noindent\textbf{Datasets.}
We evaluate our proposed ICD loss on three widely used person ReID datasets, including Market1501 \cite{Zheng2015Market}, DukeMTMC-reID \cite{Ristani2016Duke}, and MSMT17 \cite{Wei2018MSMT}. Market1501 contains 32,668 pedestrian images of 1,501 identities captured by 6 cameras. The dataset is divided into 751 training identities and 750 testing identities. DukeMTMC-reID contains 36,411 images of 1,404 identities collected from 8 cameras, including 702 identities for training and 702 identities for testing. MSMT17 is a larger person ReID dataset containing 126,441 images of 4,101 identities captured by 15 cameras. The training set contains 1,041 identities, while the testing set contains 3,060 identities. In all three datasets, the training and testing identities are non-overlapping, which is consistent with the main research motivation of our proposed ICD loss.

\noindent\textbf{Evaluation Metrics.}
Following existing person ReID methods, we use mean average precision (mAP) and cumulative matching characteristic (CMC) at Rank-1 accuracy as the evaluation metrics. The mAP measures the overall retrieval performance, while Rank-1 indicates the probability of retrieving the correct identity at the first position. Both metrics are reported in percentage (\%).

\noindent\textbf{Implementation Details.}
We use the Vision Transformer (ViT) \cite{Dosovitskiy2021ViT} as the feature extraction backbone in our experiments. The network consists of 12 Transformer encoder layers with an embedding dimension of 768. All input pedestrian images are resized to $256\times128$, and the patch size is set to $16\times16$. We initialize the backbone with parameters pre-trained on ImageNet and adopt random cropping, horizontal flipping, and random erasing for data augmentation. 
The batch size is set to 120. We optimize the network using the SGD optimizer with a momentum of 0.9 and a weight decay of $1\times10^{-4}$, and the initial learning rate is set to $8\times10^{-3}$. A cosine learning rate decay strategy is adopted, with a linear warm-up during the first 10 epochs. 
The loss weight $\lambda$ for the proposed ICD loss is set to 0.1 in all experiments.

For the implementation of the proposed ICD loss, we adopt an ArcFace-style classification layer to calculate the classification scores and the cross-entropy (CE) loss. The ICD loss is calculated based on the class-orthogonal components of the learned representations, as discussed in Section 3.2.4. 

The CE loss and ICD loss are combined to train the feature learning network, while the contribution of ICD is controlled by the loss weight $\lambda$. In the ablation studies, we also investigate the combination of ICD and triplet loss. To ensure a fair comparison, the same feature extraction backbone and training settings are maintained when comparing different loss functions under the same experimental protocol.

\begin{table}[!t]
\caption{Ablation study of the proposed ICD loss and its complementarity with Triplet and AdaSP losses on three person ReID datasets. All experiments use ViT as the backbone. Rank-1 and mAP are reported.}
\label{tab:icd_ablation}
\centering
\fontsize{9}{11}\selectfont
\setlength{\tabcolsep}{3pt}
\renewcommand{\arraystretch}{1.05}
\begin{tabular*}{\linewidth}{@{\extracolsep{\fill}}lcccccc}
\toprule
 & \multicolumn{2}{c}{Market1501} & \multicolumn{2}{c}{DukeMTMC-reID} & \multicolumn{2}{c}{MSMT17} \\
\cmidrule(lr){2-3}\cmidrule(lr){4-5}\cmidrule(lr){6-7}
Methods & R-1 & mAP & R-1 & mAP & R-1 & mAP \\
\midrule
Baseline (cross-entropy only) & 93.3 & 83.1 & 84.3 & 72.9 & 71.4 & 47.1 \\
Baseline + ICD & 94.2 & 84.5 & 84.8 & 74.1 & 72.6 & 48.2 \\
\midrule
Baseline + Triplet & 94.7 & 86.8 & 88.8 & 79.3 & 81.8 & 61.0 \\
Baseline + Triplet + ICD & \textbf{95.6} & \textbf{87.2} & \textbf{89.3} & 80.4 & \textbf{82.7} & \textbf{63.8} \\
\midrule
Baseline + AdaSP & 94.3 & 87.0 & 89.0 & 80.0 & 82.0 & 62.2 \\
Baseline + AdaSP + ICD & 95.0 & \textbf{87.2} & 89.1 & \textbf{80.9} & 82.1 & 63.0 \\
\bottomrule
\end{tabular*}
\end{table}

\subsection{Ablation Study}

In this section, we conduct ablation experiments to evaluate the effectiveness of the proposed ICD loss and its complementarity with existing loss functions. All experiments are conducted on three person ReID datasets using ViT as the feature extraction backbone. Specifically, we first adopt the cross-entropy (CE) loss as the basic training objective, and then incorporate ICD loss to evaluate its individual contribution. Furthermore, we combine ICD with two representative metric learning losses, including Triplet loss \cite{Hermans2017Triplet} and AdaSP loss \cite{Zhou2023AdaSP}, to investigate whether ICD can further improve their performance. The experimental results are shown in Table~\ref{tab:icd_ablation}.

As shown in Table~\ref{tab:icd_ablation}, incorporating ICD loss consistently improves the Rank-1 accuracy and mAP on all three datasets under different loss combinations. Compared with the CE baseline, CE + ICD achieves performance improvements on all evaluation metrics, indicating that ICD can provide additional benefits to conventional identity classification. More importantly, incorporating ICD into CE + Triplet and CE + AdaSP further improves their performance. In particular, CE + Triplet + ICD improves the mAP by 2.8\% on MSMT17, while CE + AdaSP + ICD improves the mAP by 0.9\% on DukeMTMC-reID. Among these combinations, CE + Triplet + ICD achieves the best Rank-1 accuracy on all three datasets, although CE + AdaSP + ICD obtains a slightly higher mAP on DukeMTMC-reID. Therefore, we adopt CE + Triplet + ICD as the training objective for the state-of-the-art comparison in Section 4.3. These results demonstrate that the proposed ICD loss can serve as an effective auxiliary regularization term and complement existing metric learning objectives.

\begin{table}[!t]
\caption{Comparison with representative state-of-the-art person ReID methods on Market1501, DukeMTMC-reID, and MSMT17. Rank-1 and mAP are reported. Published results are reported under the corresponding experimental settings.}
\label{tab:icd_sota}
\centering
\fontsize{9}{11}\selectfont
\setlength{\tabcolsep}{3pt}
\renewcommand{\arraystretch}{1.05}
\begin{tabular*}{\linewidth}{@{\extracolsep{\fill}}lcccccc}
\toprule
 & \multicolumn{2}{c}{Market1501} & \multicolumn{2}{c}{DukeMTMC-reID} & \multicolumn{2}{c}{MSMT17} \\
\cmidrule(lr){2-3}\cmidrule(lr){4-5}\cmidrule(lr){6-7}
Methods & R-1 & mAP & R-1 & mAP & R-1 & mAP \\
\midrule
OSNet \cite{Zhou2019OSNet} & 94.8 & 84.9 & 88.6 & 73.5 & 78.7 & 52.9 \\
SCSN \cite{Chen2020SCSN} & 95.7 & 88.5 & 91.0 & 79.0 & 83.8 & 58.5 \\
HAT \cite{Zhang2021HAT} & 95.6 & 89.5 & 90.4 & 81.4 & 82.3 & 61.2 \\
AAformer \cite{Zhu2024AAformer} & 95.4 & 87.7 & 90.1 & 80.0 & 83.6 & 63.2 \\
DCAL \cite{Zhu2022DCAL} & 94.7 & 87.5 & 89.0 & 80.1 & 83.1 & 64.0 \\
PVTReID \cite{Han2023PVTReID} & 95.3 & 87.8 & 90.0 & 80.5 & 82.3 & 63.2 \\
FusionReID \cite{Wang2025FusionReID} & 95.9 & \textbf{90.9} & 90.3 & 82.9 & 86.1 & 68.3 \\
TE-TransReID \cite{Zhang2025TETransReID} & 94.8 & 87.7 & 88.3 & 80.6 & 85.7 & 66.2 \\
ACSFormer \cite{Zhou2023ACSFormer} & 95.6 & 89.3 & 91.0 & 82.0 & 84.7 & 65.8 \\
CPL \cite{Yang2022CPL} & 95.5 & 88.8 & 89.0 & 78.8 & 81.4 & 57.8 \\
AdaSP \cite{Zhou2023AdaSP} & 95.5 & 89.8 & \textbf{91.7} & 83.0 & 85.5 & 67.1 \\
\midrule
ViT \cite{Dosovitskiy2021ViT} & 94.7 & 86.8 & 88.8 & 79.3 & 81.8 & 61.0 \\
ViT + ICD & 95.6 & 87.2 & 89.3 & 80.4 & 82.7 & 63.8 \\
\midrule
TransReID \cite{He2021TransReID} & 95.2 & 89.5 & 90.7 & 82.6 & 86.2 & 69.4 \\
TransReID + ICD & \textbf{96.5} & 89.8 & 91.0 & \textbf{84.0} & \textbf{86.7} & \textbf{69.5} \\
\bottomrule
\end{tabular*}
\end{table}

\subsection{Comparison with State-of-the-Art Methods}

In this section, we compare our proposed ICD loss with several representative state-of-the-art person ReID methods on three datasets. The experimental results are shown in Table 2. These methods include CNN-based and Transformer-based methods, as well as methods that introduce additional loss functions, such as CPL and AdaSP. In our experiments, we incorporate ICD loss into two representative feature learning networks, ViT and TransReID, using the combination of CE, triplet, and ICD losses as the training objective. As shown in Table 2, incorporating ICD consistently improves both Rank-1 accuracy and mAP on all three datasets. Specifically, ViT + ICD improves mAP by 2.8\% on MSMT17, while TransReID + ICD improves Rank-1 accuracy by 1.3\% on Market1501 and mAP by 1.4\% on DukeMTMC-reID.

Compared with existing state-of-the-art methods, our TransReID + ICD achieves competitive performance across the three datasets. In particular, it obtains the highest Rank-1 accuracy on Market1501 and MSMT17 among the compared methods. Compared with CPL and AdaSP, our method achieves higher mAP on DukeMTMC-reID and MSMT17. Although these methods adopt different backbone networks and training settings, the consistent improvements obtained on both ViT and TransReID demonstrate the effectiveness of the proposed ICD loss. 

\begin{figure}
	\centering
		\includegraphics[width=0.85\textwidth]{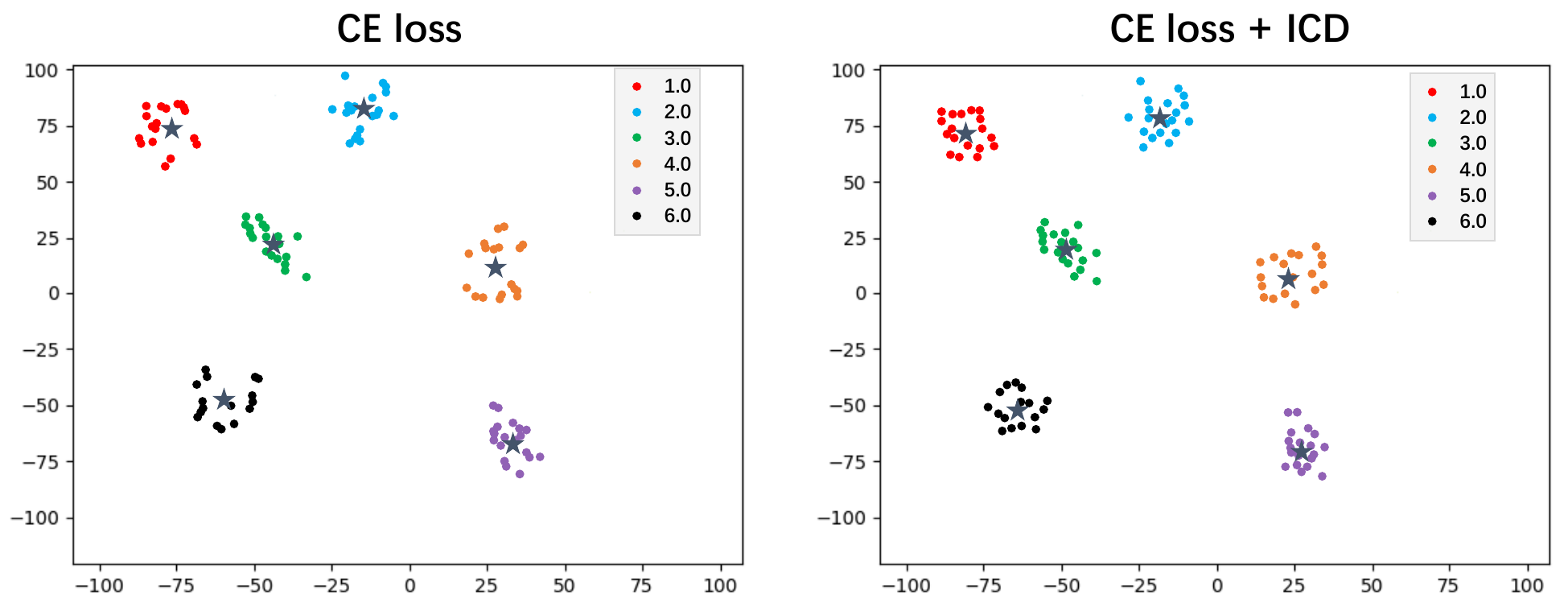}
\caption{t-SNE visualization of the learned representations of six randomly selected training identities from Market1501, with 16 images per identity. Different colors denote different identities, and the blue stars represent their corresponding class-weight vectors. The left and right figures show the feature distributions learned with CE loss and CE + ICD loss, respectively.}
	\label{fig:icd_visualization}
\end{figure}

\subsection{Visualization}

To further analyze the effectiveness of our proposed ICD loss, we use t-SNE to visualize the learned pedestrian representations on Market1501. Specifically, we randomly select six training identities, with 16 images for each identity, and compare their feature distributions under two different training objectives, including CE loss and CE + ICD loss. As shown in Figure~\ref{fig:icd_visualization}, different colors represent different pedestrian identities, while the blue stars denote their corresponding class-weight vectors.

It can be observed from Figure~\ref{fig:icd_visualization} that, when only CE loss is used, the learned representations of the same identity are distributed irregularly around the corresponding class-weight vector. After incorporating ICD loss, these representations tend to be more evenly distributed around their class-weight vectors, forming more circular distributions.  Meanwhile, the representations of different identities remain well separated, without an obvious increase in their distances from the corresponding class-weight vectors in the visualization. These results are consistent with the geometric intuition illustrated in Figure 1(c). Instead of simply pushing the learned representations away from their class-weight vectors, the proposed ICD loss aims to maximize intra-class differences in classification-irrelevant directions, while maintaining identity discrimination.

\FloatBarrier
\section{Conclusion}

In this paper, we propose an intra-class diversity (ICD) loss for person re-identification. Different from existing loss functions that mainly focus on intra-class compactness and inter-class separation, our proposed ICD loss aims to maximize intra-class differences in classification-irrelevant directions. Specifically, based on the geometric relationship between the learned representation vector and its corresponding class-weight vector, we propose to maximize the differences among representations of the same identity in the class-orthogonal directions. In this way, ICD provides an additional regularization objective for preserving intra-class diversity without modifying the original feature extraction network. Extensive experiments on three widely used person ReID datasets demonstrate the effectiveness of the proposed ICD loss. Furthermore, the experimental results show that ICD can complement existing metric learning losses and further improve ReID performance.

\end{document}